\documentclass[letterpaper, 10 pt, conference]{ieeeconf}

\IEEEoverridecommandlockouts %

\usepackage{amsmath,amsfonts,amssymb}
\usepackage{amsfonts} %
\usepackage{mathrsfs} %
\usepackage{bbold} %
\usepackage{stix} %
\usepackage{pgfplots}
\usepackage[ruled, vlined, linesnumbered]{algorithm2e}
\usepackage{subfig}
\usepackage{tabularx}
\usepackage{booktabs} %
\usepackage{multirow}
\usepackage{siunitx}
\usepackage[export]{adjustbox}

\usepackage{caption}
\usepackage{csquotes}

\usepackage{graphicx}
\graphicspath{{figures/}}
\usepackage{transparent}

\usepackage{xargs} %

\usepackage[style=ieee, backend=biber, bibencoding=utf8, maxbibnames=3, mincitenames=1, maxcitenames=2, url=false, doi=false, isbn=false, citestyle=numeric-comp, natbib=true, eprint=true]{biblatex}

\usepackage[colorinlistoftodos,prependcaption,textsize=small]{todonotes}

\newcommandx{\change}[2][1=]{\todo[inline,linecolor=blue,backgroundcolor=blue!25,bordercolor=blue,#1]{#2}}

\usepackage{bm} %
\usepackage{lipsum}

\usepackage{scalerel} %

\pgfplotsset{compat=newest}
\pgfplotsset{plot coordinates/math parser=false}
\usetikzlibrary{decorations.markings, positioning}

\newlength\figureheight 
\newlength\figurewidth 

\makeatletter
\edef\@figdir{_pgfcache}

\catcode`\#=11
\catcode`\|=6
\newcommand{\@basicpgfpreamble}[1]{%
    \unexpanded{%
        \documentclass{standalone}^^J
        \usepackage{pgf}^^J
        \RequirePackage{luatex85}^^J
        \let\oldpgfimage\pgfimage^^J
        \renewcommand{\pgfimage}[2][]{\oldpgfimage[#1]{|1/#2}}^^J
    }%
}
\catcode`\#=6
\catcode`\|=12

\let\@pgfpreamble\@basicpgfpreamble

\newcommand{\setpgfpreamble}[1]{%
    \renewcommand{\@pgfpreamble}[1]{\@basicpgfpreamble{##1}\unexpanded{#1}}
}

\newcounter{@pgfcounter}
\newwrite\@pgfout
\newread\@pgfin

\newcommand{\importpgf}[3][]{%
    \IfFileExists{#2/#3}{}{\errmessage{importpgf: File #2/#3 not found}}%
    \edef\@figfile{\jobname-\the@pgfcounter}%
    \providecommand{\@writetempfile}{}%
    \renewcommand{\@writetempfile}[1]%
    {%
        \immediate\openout\@pgfout=##1%
        \immediate\write\@pgfout{\@pgfpreamble{#2}}%
        \immediate\write\@pgfout{\string\begin{document}}%
        \immediate\openin\@pgfin=#2/#3%
        \begingroup\endlinechar=-1%
            \loop\unless\ifeof\@pgfin%
                \readline\@pgfin to \@fileline%
                \ifx\@fileline\@empty\else%
                    \immediate\write\@pgfout{\@fileline}%
                \fi%
            \repeat%
        \endgroup%
        \immediate\closein\@pgfin%
        \immediate\write\@pgfout{\string\end{document}}%
        \immediate\closeout\@pgfout%
    }%
    \def\@compile%
    {%
        \immediate\write18{lualatex -interaction=batchmode -output-directory="\@figdir" \@figdir/\@figfile.tex}%
    }%
    \IfFileExists{\@figdir/\@figfile.pdf}%
    {%
        \@writetempfile{\@figdir/tmp.tex}%
        \edef\@hashold{\pdfmdfivesum file {\@figdir/\@figfile.tex}}%
        \edef\@hashnew{\pdfmdfivesum file {\@figdir/tmp.tex}}%
        \ifnum\pdfstrcmp{\@hashold}{\@hashnew}=0%
            \relax%
        \else%
            \@writetempfile{\@figdir/\@figfile.tex}%
            \@compile%
        \fi%
    }%
    {%
        \@writetempfile{\@figdir/\@figfile.tex}%
        \@compile%
    }%
    \IfFileExists{\@figdir/\@figfile.pdf}%
    {\includegraphics[#1]{\@figdir/\@figfile.pdf}}%
    {\errmessage{Error during compilation of figure #2/#3}}%
    \stepcounter{@pgfcounter}%
}
\makeatother

\setpgfpreamble{
}

\newcommand{\plotpath}{./../results/figures}

\title{\LARGE \bf Modeling and Testing Multi-Agent Traffic Rules within Interactive Behavior Planning}
\author{Klemens Esterle$^{1}$, Luis Gressenbuch$^{2}$, and Alois Knoll$^{2}$%
	\thanks{$^{1}$Klemens Esterle is with fortiss GmbH, Research Institute of the Free State of Bavaria, Munich, Germany, esterle@fortiss.org}%
	\thanks{$^{2}$Luis Gressenbuch and Alois Knoll are with Robotics, Artificial Intelligence and Real-time Systems, Technische Universit\"{a}t M\"{u}nchen, Munich, Germany}%
}

\begin{document}

\maketitle
\thispagestyle{empty}
\pagestyle{empty}

\global\csname @topnum\endcsname 0
\global\csname @botnum\endcsname 0

\newcommand{\figurename}{Fig. }

\newcommand {\vect} {\boldsymbol}
\newcommand {\matr} {\boldsymbol}

\newcommand{\state} {\vect{x}}
\newcommand{\stateSpace} {\vect{\mathcal{X}}}
\newcommand{\beliefstate} {\vect{b}}
\newcommand{\contr} {\vect{u}}
\newcommand{\contrSpace} {\vect{\mathcal{U}}}
\newcommand{\meas} {\vect{y}}
\newcommand{\procNoise}{\vect{w}}

\newcommand {\cov}  {\matr{\Sigma}}

\newcommand{\stateNoDelta}{\hat\state}
\newcommand{\contrNoDelta}{\hat\contr}
\newcommand{\procNoiseNoDelta}{\hat\procNoise}

\newcommand{\abc}[2][\empty]{%
  \ifthenelse{\equal{#1}{\empty}}
    {no opt, mand.: \textbf{#2}}
    {opt: \textbf{#1}, mand.: \textbf{#2}}
}

\newcommand {\noiseu} {\procNoise}
\newcommand {\covu} {\matr{\Sigma_{\noiseu,}}}

\newcommand {\noiseuNoDelta} {\procNoiseNoDelta}
\newcommand {\covuNoDelta} {\matr{\Sigma_{\noiseuNoDelta,}}}

\newcommand {\defnoiseu}[1][\empty]{
 \ifthenelse{\equal{#1}{\empty}}
    {\noiseu\sim N(0,\covu)}
    {\noiseu_{#1}\sim N(0,\covu_{#1})}
}

\newcommand {\defnoiseuNoDelta}[1][\empty]{
 \ifthenelse{\equal{#1}{\empty}}
    {\noiseuNoDelta\sim N(0,\covuNoDelta)}
    {\noiseuNoDelta_{#1}\sim N(0,\covuNoDelta_{#1})}
}

\newcommand {\covm} {\matr{R}}
\newcommand {\noisem} {\vect{\nu}}
\newcommand {\defnoisem}[1][\empty]{
 \ifthenelse{\equal{#1}{\empty}}
    {\noisem\sim N(0,\covm)}
    {\noisem_{#1}\sim N(0,\covm_{#1})}
}

\newcommand{\stateB}{\vect{\xi}}
\newcommand{\contrB}{\vect{\nu}}
\newcommand{\procNoiseB}{\vect{\omega}}

\newcommand{\AB}{\mathcal{A}}
\newcommand{\BB}{\mathcal{B}}
\newcommand{\WB}{\mathcal{W}}
\newcommand{\costStateB}{\mathcal{Q}}
\newcommand{\costContrB}{\mathcal{R}}
\newcommand{\covStatesB}{\mathcal{S}_{\state}}
\newcommand{\covProcNoiseB}{\mathcal{S}_{\procNoise}}

\newcommand{\FB}{\mathcal{F}}

\newcommand{\cct}{\vect{t}}
\newcommand{\ccsval}{s}
\newcommand{\ccT}{\matr{T}}
\newcommand{\ccsvec}{\vect{s}}

\newcommand{\costState}{\matr{Q}}
\newcommand{\costContr}{\matr{R}}
\newcommand{\feedbackMatrix}{\matr{K}}
\newcommand{\cost}{J}

\newcommand{\stateConstraintMatrix}{\matr{C}}
\newcommand{\stateConstraintVector}{\vect{c}}

\newcommand{\stateConstraintFunc}{c}

\newcommand{\contrConstraintMatrix}{\matr{D}}
\newcommand{\contrConstraintVector}{\vect{d}}

\newcommand{\contrConstraintFunc}{d}

\newcommand{\stateRef}{\state^{*}}
\newcommand{\contrRef}{\contr^{*}}

\newcommand{\stateDelta}{\Delta\state}
\newcommand{\contrDelta}{\Delta\contr}

\newcommand {\Comment}[1]{\textcolor{blue}{#1}}

\newcommand {\partialder}[4][\bigg]{\frac{\partial #2}{\partial #3}#1|_{#4}}
\newcommand {\partialdernoarg}[3][\bigg]{\frac{\partial #2}{\partial #3}#1}

\newcommand{\nat}{\mathbb{N}}
\newcommand{\real}{\mathbb{R}}
\newcommand{\compl}{\mathbb{C}}

\newcommand{\norm}[1]{\left\| #1 \right\|}

\newcommand{\half}{\frac{1}{2}}

\newcommand{\parenth}[1]{ \left( #1 \right) }
\newcommand{\bracket}[1]{ \left[ #1 \right] }
\newcommand{\accolade}[1]{ \left\{ #1 \right\} }
\newcommand{\pardevS}[2]{ \delta_{#1} f(#2) }
\newcommand{\pardevF}[2]{ \frac{\partial #1}{\partial #2} }

\newcommand{\vecii}[2]{\begin{pmatrix} #1 \\ #2 \end{pmatrix}}
\newcommand{\veciii}[3]{\begin{pmatrix}  #1 \\ #2 \\ #3	\end{pmatrix} }
\newcommand{\veciv}[4]{\begin{pmatrix}  #1 \\ #2 \\ #3 \\ #4	\end{pmatrix}}

\newcommand{\matii}[4]{\left[ \begin{array}[h]{cc} #1 & #2 \\ #3 & #4 \end{array} \right]}
\newcommand{\matiii}[9]{\left[ \begin{array}[h]{ccc} #1 & #2 & #3 \\ #4 & #5 & #6 \\ #7 & #8 & #9	\end{array} \right]}

\newcommand{\transp}{^{\intercal}}
\newcommand{\Reg}{$^{\textregistered}$}
\newcommand{\reg}{$^{\textregistered}$ }
\newcommand{\Tm}{\texttrademark}
\newcommand{\tm}{\texttrademark~}
\newcommand {\bsl} {$\backslash$}

\newtheorem{theorem}{Theorem}[section]
\newtheorem{lemma}[theorem]{Lemma}
\newtheorem{corollary}[theorem]{Corollary}
\newtheorem{remark}[theorem]{Remark}
\newtheorem{definition}[theorem]{Definition}
\newtheorem{equat}[theorem]{Equation}
\newtheorem{example}[theorem]{Example}
\newcommand{\insertfigure}[4]{ %
	\begin{figure}[htbp]
		\begin{center}
			\includegraphics[width=#4\textwidth]{#1}
		\end{center}
		\vspace{-0.4cm}
		\caption{#2}
		\label{#3}
	\end{figure}
}

\newcommand{\refFigure}[1]{\figurename \ref{#1}}
\newcommand{\refChapter}[1]{chapter \ref{#1}}
\newcommand{\refSection}[1]{Section \ref{#1}}
\newcommand{\refParagraph}[1]{paragraph \ref{#1}}
\newcommand{\refEquation}[1]{(\ref{#1})}
\newcommand{\refTable}[1]{Table \ref{#1}}
\newcommand{\refAlgorithm}[1]{Algorithm \ref{#1}}

\newcommand{\rigidTransform}[2]
{
	${}^{#2}\!\mathbf{H}_{#1}$
}

\newcommand{\code}[1]
 {\texttt{#1}}

\newcommand{\comment}[1]{\marginpar{\raggedright \noindent \footnotesize {\sl #1} }}

\newcommand{\clearemptydoublepage}{%
  \ifthenelse{\boolean{@twoside}}{\newpage{\pagestyle{empty}\cleardoublepage}}%
  {\clearpage}}

\newcommand{\etAl}{\emph{et al.}\mbox{ }}

\newcommand{\todoi}[1]{\todo[inline]{#1}}

\newcommand{\SymbolPlaceholder}{\scaleobj{0.8}{\blacksquare}}
\newcommand \Global {\scaleobj{0.95}{\square}}
\newcommand \Next       {\scaleobj{0.85}{\bigcirc}}
\newcommand \Final       {\lozenge}
\newcommand \Or {\vee}
\newcommand \lUntil      {\mathsf{U}} %
\newcommand \ImpliesCustomized {\Rightarrow}

\newcommand \AtomicProposition {\pi}
\newcommand \AtomicPropositionSet {\Pi}

\newcommand \finiteLTL {\text{LTL}_f}

\newcommand \LTLsymbol {\sigma}
\newcommand \LTLword {w}
\newcommand \LTLformula {\varphi}
\newcommand \LTLFormulafinite {p}
\newcommand \LTLfalive {\small{\textsf{\textit{alive}}}}
\newcommand \true {\mathbf{T}}

\newcommand \dfa {\Lambda}
\newcommand \AutomatonStateSet {Q}
\newcommand \AutomatonState {q}
\newcommand \AutomatonInitialState {q_{0}}
\newcommand \AutomatonAlphabet {\Sigma}
\newcommand \AutomatonTransitionFunction {\delta}
\newcommand \AutomatonAcceptingStates {F}
\newcommand \AutomatonWeightingFunction {W}

\newcommand \LabeledMarkovStructure {\mathcal{M}}
\newcommand \LabeledMarkovAPSet {\Pi}
\newcommand \LabeledMarkovLabelingFunction {\mathcal{L}}

\newcommand{\productstate}{z}
\newcommand {\weight} {\omega}
\newcommand \transpose {\mathsf{T}}

\newcommand{\labelstyle}[1]{\small{\textsf{\textit{{#1}}}}}
\newcommand \labelInDirectFront {\labelstyle{in-direct-front}}
\newcommand \labelRight {\labelstyle{right}}
\newcommand \labelLeft {\labelstyle{left}}
\newcommand \labelFront {\labelstyle{in-front}}
\newcommand \labelBehind {\labelstyle{behind}}
\newcommand \labelMerged {\labelstyle{merged}}
\newcommand \labelCongested {\labelstyle{congested}}
\newcommand \labelSafeDistanceFront {\labelstyle{sd-front}}
\newcommand \labelColliding {\labelstyle{collide}}

\newcommand{\labelEgoAgent}{i}
\newcommand{\labelInRelationToAgent}{j}
\newcommand{\inrelationto}{\rightarrowtail}

\newtheorem{traffic-rule}{Traffic Rule}

\newcommand \Rulebook {\mathcal{R}}

\newcommand {\SetOfAgents} {\mathcal{P}}
\newcommand {\Agent} {P}
\newcommand {\NumOfAgents} {N}
\newcommand {\NumOfTimesteps} {K}
\newcommand {\NumOfRules} {m}
\newcommand{\Agenti}[1][]{{{#1}}}
\newcommand {\StateSpace} {\mathcal{S}}
\newcommand {\StateSpacei}[1][]{{\StateSpace_{#1}}}
\newcommand {\MarkovState} {s}
\newcommand {\VehicleState} {{x}}
\newcommand {\VehicleStateSpacei}[1][]{{X}_{#1}}
\newcommand {\ActionSpace} {{A}}
\newcommand {\JointActionSpace} {\mathcal{A}}
\newcommand {\Action} {a}
\newcommand {\JointAction} {\boldsymbol{a}}
\newcommand {\ActionSpacei}[1][]{{\ActionSpace_{#1}}}
\newcommand {\TransitionFunction} {T}
\newcommand {\RewardFunction} {R}
\newcommand {\reward} {r}
\newcommand {\rewardVector} {\vect{\reward}}
\newcommand {\DiscountFactor} {\gamma}

\newcommand{\letlo}{\preceq_\mathrm{TLO}}
\newcommand{\thesholdtlo}{\tau}
\newcommand{\SubscriptCol}{\mathrm{col}}
\newcommand{\SubscriptZip}{\mathrm{zip}}
\newcommand{\SubscriptBase}{\mathrm{base}}
\newcommand{\SubscriptSD}{\mathrm{sd}}

\newcommand{\classnamestyle}[1]{\small{\textsf{{#1}}}}

\newcommand{\behaviorOthers}{\mathcal{B}_{\mathrm{o}}}
\newcommand{\behaviorModelUnterTest}{\mathcal{B}_{\mathrm{e}}}

\begin{abstract}
Autonomous vehicles need to abide by the same rules that humans follow. Some of these traffic rules may depend on multiple agents or time.
Especially in situations with traffic participants that interact densely, the interactions with other agents need to be accounted for during planning.
To study how multi-agent and time-dependent traffic rules shall be modeled, a framework is needed that restricts the behavior to rule-conformant actions during planning, and that can eventually evaluate the satisfaction of these rules.
This work presents a method to model the conformance to traffic rules for interactive behavior planning and to test the ramifications of the traffic rule formulations on metrics such as collision, progress, or rule violations. The interactive behavior planning problem is formulated as a dynamic game and solved using Monte Carlo Tree Search, for which we contribute a new method to integrate history-dependent traffic rules into a decision tree. To study the effect of the rules, we treat it as a multi-objective problem and apply a relaxed lexicographical ordering to the vectorized rewards.
We demonstrate our approach in a merging scenario. We evaluate the effect of modeling and combining traffic rules to the eventual compliance in simulation. 
We show that with our approach, interactive behavior planning while satisfying even complex traffic rules can be achieved. Moving forward, this gives us a generic framework to formalize traffic rules for autonomous vehicles.
\end{abstract}

\IEEEpeerreviewmaketitle

\section{Introduction}
\label{sec:introduction}

Traffic rules have been created to help humans manage the otherwise chaotic traffic environment. When sharing the road with human drivers, autonomous vehicles will have to obey the same rules humans do, which often depend on the actions of other agents or the past.
Previous work has proposed a method for combining model checking techniques for traffic rule satisfaction with motion planning in static environments \cite{ReyesCastro2013}. However, dense scenarios have shown to be difficult for such motion planning approaches, as they do not model the interactions with human drivers and are thus unable to correctly anticipate human reactions. The research line of \textit{interactive behavior planning} addresses this problem but has mostly ignored the aspect to obey traffic rules other than collision prevention and speed compliance.
The problem setting that is closest to ours is \cite{Chaudhari2014a}. However, the approach is limited to rules that depend on only one agent and consequently cannot incorporate rules such as keeping a safe distance or merging in a zipper fashion.

Merging scenarios have proven to be challenging due to the dense interaction with others, combined with the eventual lane ending, prompting the need for the driver to make a decision. Game-theoretic approaches offer an elegant way to model such interactions. We propose a game-based planning approach that monitors traffic rules, which depend on multiple agents and past information, at runtime. We refer to these rules by \textit{multi-agent, time-dependent traffic rules}. To solve this formulation, we use Monte Carlo Tree Search (MCTS). 
Evaluating time-dependent traffic rules violates the Markov property, i.e., it does not rely only on the current state but also on previous states. This raises the question how to integrate those rules in MCTS, which depends on the Markov property to be computationally efficient.

Specifically, we contribute a game-based planning method monitoring multi-agent and time-dependent traffic rules,
and a method to model non-Markovian traffic rules within MCTS.
To study the effect of modeling a certain traffic rule within a set of rules, we treat the set of rules to be priority-ordered, and incorporate it to MCTS by leveraging methods from multi-objective optimization theory.
In simulations based on real-world data, we use the same runtime monitors for both planning and evaluating a scenario run. We provide a comparable study of the ramifications on collision, progress, and the violation of rules in the simulated scenarios when modeling the applicable traffic rules within MCTS.

\section{Related Work}
\label{sec:related_work}

To safely drive in mixed traffic, multiple goals, such as collision and safety metrics, various traffic rules, and comfort metrics should be considered. Some of these goals are strictly preferred over others. Expressing the preference using a weighted sum scalarization is tedious, sometimes even impossible.
Therefore, \citet{ReyesCastro2013} define the preference relation in lexicographical order. %
Compared to a weighted sum scalarization, their method does not require to adapt the weights of all the cost terms. 

\citet{ReyesCastro2013} propose a sampling-based method to generate a trajectory that minimizes a set of formalized traffic rules in a finite fragment of Linear Temporal Logic (LTL).
They only check collisions with static obstacles, as their use case is limited to a static environment while traveling at a constant speed.
Their state space consists of position and orientation, which yields an analytic steering function, and thus enables the efficient connection of samples in the tree. However, there is no straight forward way to include dynamic obstacles from a prediction module.
When LTL is employed to formalize the traffic rules within motion planning \cite{ReyesCastro2013, Esterle2019a}, violating such a formula only yields a binary satisfaction signal. With these planning approaches, neither the reactions of others are taken into account, nor does the violation penalty incorporate any prediction uncertainty, that circumvents this lack of interaction.

Satisfying multi-agent traffic rules has been realized by establishing contracts between vehicles \cite{Decastro2018}. However, their approach is used for deriving requirements during the design phase, not for penalizing or restricting actions during planning.

\citet{Chaudhari2014a} define a two-player non-zero-sum non-cooperative game. Both the ego agent as well as the environment (the other agents) try not to violate any traffic rules, which are expressed in LTL. Each agent builds up a tree as in \cite{ReyesCastro2013}. That prohibits them from incorporating rules that depend on other agents than the ego agent, such as the safe distance or zipper merge rule.

\citet{Lee2019} perform runtime verification on full traces of MCTS. They formalize simple single-agent traffic rules in LTL, and demonstrate it for an intersection. Instead, our approach can incorperate multi-agent and time-dependent traffic rules in an efficient way, by exploiting automata-based model checking in the search tree.

To summarize, no work currently exists that allows to study multi-agent traffic rules in dense scenarios by modeling it as part of the interactive planning problem.

\section{Problem Statement}
\label{sec:problem}
In a lane-merging scenario populated with multiple agents, this work aims to plan the behavior of a single agent, i.e. the generation of a sequence of desired future dynamic states $\{\VehicleState(t=t_1), \VehicleState(t=t_2), ..., \VehicleState(t=t_\NumOfTimesteps)\}$ while obeying the applying traffic rules.
We assume perfect observation of the dynamic state of the other agents, of the map and our localization within that.

We treat this setting of interactive behavior planning as a dynamic game, that formally consists of:
\begin{itemize}
	\item A set of $\NumOfAgents$ agents, each having a dynamic state $\VehicleState_i \in \VehicleStateSpacei[i]$,
	\item An environment state $\MarkovState \in \StateSpacei = \times \VehicleStateSpacei[i]$ with state space $\StateSpacei$,
	\item Agent \num{1} (referred to as \blockquote{ego agent}) having an action space $\ActionSpacei[1]$ with discrete actions,%
	\item Agents $2 ... \NumOfAgents$ (referred to as \blockquote{other agents}) following a behavior model $\behaviorOthers$, that determines the agent's action $\Action_i^t \sim \behaviorOthers(\MarkovState^t)$.
\end{itemize}

Based on the joint action of the agents, the environment described by joint state $\MarkovState$ transitions to the next state $\MarkovState'$. The ego agent gets a reward $\reward(\MarkovState, \MarkovState', \JointAction)$ after joint action $\JointAction \in \JointActionSpace = \times \ActionSpacei[i]$ is applied. 
The goal is to find an optimal action $\Action$ for the ego vehicle that maximizes it's cumulative reward $\sum_{0}^{\NumOfTimesteps} \DiscountFactor^k \reward_k$ along a planning horizon of $\NumOfTimesteps$ steps, where the reward incorporates penalties for discomfort, traffic rule violation, and collision. The discount factor is denoted by $\DiscountFactor$.
As an optimal solution of the decision problem is infeasible due to the size of the state space, we will use MCTS to obtain an approximation of the optimal solution using sampling.

We will use the formalized traffic rules from \cite{Esterle2020b}, which are formulated in LTL on finite traces ($\finiteLTL$).
Some of these traffic rules, such as the zipper merge or overtaking, do not rely only on the current state $\MarkovState$ but also on previous information. To be used in an interactive planning framework as described above, we aim to find a Markovian formulation for evaluating the history-dependent rules for each agent.

\section{Preliminaries}
\label{sec:preliminaries}

\subsection{Linear Temporal Logic on Finite Traces}
\label{subsec:ltl}
Linear Temporal Logic is a discrete formal logic to reason not just about an absolute truth but about truths which might hold only at some points in time. It can thus be used to represent non-Markovian properties.

Let $\LabeledMarkovAPSet$ be a set of atomic propositions. The powerset of $\LabeledMarkovAPSet$, i.e., the set of all subsets of $\LabeledMarkovAPSet$, is denoted by $2^\Pi$. A labeling function $\LabeledMarkovLabelingFunction: \StateSpace \rightarrow 2^\Pi$ obtains the labels from state $\MarkovState$.

Formally, the language $\LTLformula$ of LTL formulas is defined as
\begin{align*}
\LTLformula ::= &\, \AtomicProposition\, | \, \lnot \LTLformula \, | \, \LTLformula_1 \land \LTLformula_2 \, | \LTLformula_1 \lor \LTLformula_2  | \LTLformula_1 \ImpliesCustomized \LTLformula_2 \, | \, \Next \LTLformula \, | \\ 
&\, \LTLformula_1 \lUntil \LTLformula_2 \, | \, \Global \LTLformula \, | \, \Final \LTLformula,
\nonumber
\end{align*}
where $\AtomicProposition \in \AtomicPropositionSet$ denotes an \emph{atomic} proposition, $\lnot$ (resp. $\land$, $\lor$, $\ImpliesCustomized$) denote the boolean operators ``not'', ``and'', ``or'' and ``implies'',
and $\Next$, (resp. $\lUntil$, $\Global$, $\Final$) denote the temporal operators ``next'', ``until'', ``globally'' (or ``always''), ``finally'' (or ``eventually'').
See \cite{Baier2008} for a definition of the semantics.

In the context of continuously replanning over a receding horizon, we use $\finiteLTL$, as it provides a formalism to reason over bounded periods of time. It uses the same syntax as LTL.
We refer to the work of \citet{DeGiacomo2013} for definitions of the semantics of $\finiteLTL$. 

Following the categorization of \citet{Manna1990}, we will restrict ourselves to formulas with obligation properties, which include safety properties as well as guarantee properties. 
Safety formulas can be represented as $\Global \LTLFormulafinite$, whereas guarantees are generally captured by $\Final \LTLFormulafinite$, where $\LTLFormulafinite$ is a finite LTL formula only consisting of atomic propositions and the operators $\lnot, \lor, \land, \ImpliesCustomized$, $\Next$, and past-LTL operators.

\subsection{Automaton-based Verification of $\finiteLTL$ Formulas}
To verify if a trace satisfies the $\finiteLTL$ formula, we utilize automata-based model checking. Given a formula in $\finiteLTL$, a deterministic finite automaton (DFA) can be constructed, that recognizes words satisfying the formula. For the $\finiteLTL$ formula to be translated into a DFA, we follow the concept of \cite{DeGiacomo2013}. An additional atomic proposition $\LTLfalive$ is introduced to the formula before translation to the automaton. This symbol will be set to true as long as the search horizon has been reached, otherwise false. The automaton is defined over the alphabet $\AutomatonAlphabet = \AtomicPropositionSet$, i.e., the set of atomic propositions of the $\finiteLTL$ formula.

\begin{definition}
	A deterministic finite automaton is a tuple $\dfa = (\AutomatonStateSet, \AutomatonInitialState, \AutomatonAlphabet, \AutomatonTransitionFunction, \AutomatonAcceptingStates)$, where
	\begin{itemize}
		\item $\AutomatonStateSet$ is a set of states
		\item $\AutomatonInitialState \in \AutomatonStateSet$ is the initial state
		\item $\AutomatonAlphabet$ is a finite alphabet
		\item $\AutomatonTransitionFunction : \AutomatonStateSet \times \AutomatonAlphabet \rightarrow \AutomatonStateSet $ is a transition function
		\item $\AutomatonAcceptingStates \subseteq \AutomatonStateSet$ denotes a set of accepting (or final) states.
	\end{itemize}
\end{definition}

For a word $\LTLword$, given as a sequence of symbols $\LTLsymbol \in \AutomatonAlphabet$ and a DFA $\dfa$, the automaton state is sequentially updated from the initial automaton state with respect to $\delta$ and $\LTLword$. Therefore, if the automaton halts in an accepting state, the trajectory satisfies the formula.

\subsection{Monte Carlo Tree Search for Behavior Planning}
MCTS is an online method to approximate the solution of sequential decision problems via sampling.
Throughout the search, the estimate of the state-action-value function $Q(\MarkovState, \Action)$, which maps the expected cumulative reward of performing action $\Action$ in state $\MarkovState$, is updated iteratively.
Each iteration consists of \textit{selection}, \textit{expansion}, \textit{rollout}, and \textit{backup}.

For \textit{selecting} a new node, the tree is traversed following the tree policy of best actions until a state with untried actions is reached. 
A commonly used selection strategy is called Upper Confidence Bounds for Trees (UCT), which balances exploitation and exploration.
During \textit{expansion}, an untried actions is applied, and the child node is added to the tree.
From the newly expanded node, a value estimate is obtained using a heuristic \textit{rollout} (also called \textit{simulation}): Based on a default policy, actions are applied until a terminal state is reached. The simplest case is to make uniform random moves.
The observed value is \textit{backed up} to the root node and used to update $Q(\MarkovState, \Action)$.
The search is repeated until some predefined computation budget (time or iterations) is reached. Finally, the best performing root action is returned.

\citet{Lenz2016} applied this concept to cooperative driving, as they introduce cooperative costs that include the costs of all agents.
When sampling the actions of all agents, the size of the search tree grows exponentially with the number of agents. To mitigate that, only a subset of all agents are included in the joint action, while the actions of the remaining agents are based on a predefined model. 
We will base our work on a special case of this, called \textit{Single-Agent MCTS}, where all other agents are predicted using this model. However, the approach can be easily extended to multiple agents.

\section{Approach}
\label{sec:solution}
In this section, we first present our method for monitoring traffic rules within interactive behavior planning based on MCTS. We describe how we obtain a reward from the rule violation. We then describe how to extend MCTS to rewards in a lexicographical ordering.

\subsection{Runtime Monitoring within Monte Carlo Tree Search}
To allow for a valuation of the traffic rules given in $\finiteLTL$, we label the environment state $\MarkovState$ according to whether an atomic proposition such as \blockquote{agent $i$ is in lane} or \blockquote{agent $i$ is in front of $j$} is \textit{true} or \textit{false}. 
A description of the necessary labels to express the traffic rules is available in \cite{Esterle2020b}.

To monitor temporal rules within MCTS, the straightforward way is to evaluate the run from the root node to the current node for each expansion and simulation step.
However, this will significantly limit the performance of the MCTS. Instead, we exploit the structure of the decision tree, by augmenting the tree nodes of the MCTS by the automaton state. Thus, non-Markovian properties captured in LTL can be efficiently verified during the MCTS by encoding necessary historic information into the automaton's state.

We first translate each $\LTLformula_i$ formula into their corresponding DFA representation $\dfa_i$. Instead of combining them to a single product automaton, we consider them as independent automata, which is advantageous for the time complexity of finding a valid transition. The states of $\NumOfRules$ automata form the automaton state vector $\overline{\AutomatonState}$

\begin{equation}
 \overline{\AutomatonState} = 
	 \begin{pmatrix}
		\AutomatonState_1 & \hdots & \AutomatonState_\NumOfRules
	 \end{pmatrix}^\transpose.
\end{equation}

\begin{figure}[tb]
	\vspace{0.15cm}
	\centering
	\scriptsize
	\def\svgwidth{\columnwidth}
	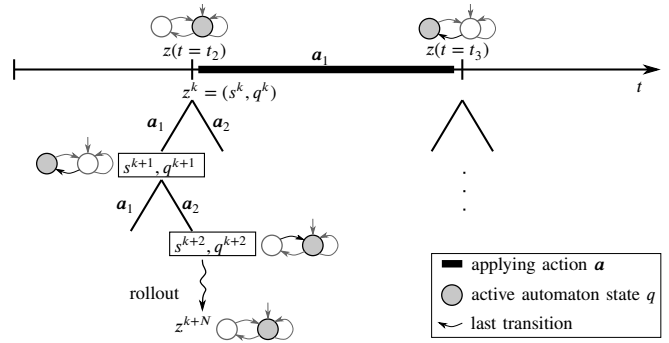
	\caption{The evolution of the product state $\productstate$ for a single rule monitor. An MCTS planning step is depicted exemplary at $t_2$.}
	\label{fig:mcts_over_time}
	\vspace{-0.15cm}
\end{figure}

With the joint state $\MarkovState$ and the automaton state vector $\overline{\AutomatonState}$, we can define the combined state $\productstate$

\begin{equation}
\productstate^k = 
\begin{pmatrix}
\MarkovState^k & \overline{\AutomatonState}^k
\end{pmatrix}^\transpose.
\end{equation}
The successor state is then defined by
\begin{equation}
	\productstate^{k+1} = 
		\begin{pmatrix} 
			\MarkovState^{k+1} \\ 
			\AutomatonTransitionFunction_{1} \big( \AutomatonState_{1}^{k}, \LabeledMarkovLabelingFunction(\MarkovState^{k+1}) \big) \\
			\vdots \\
			\AutomatonTransitionFunction_{\NumOfRules} \big( \AutomatonState_{\NumOfRules}^{k}, \LabeledMarkovLabelingFunction(\MarkovState^{k+1}) \big) \\
		\end{pmatrix}
\end{equation}
where $\MarkovState^{k+1}$ is defined by the joint action $\JointAction$.
\refFigure{fig:mcts_over_time} illustrates the evolution of the product state $\productstate$ during MCTS and over time for a single rule monitor.

\subsection{Costs for Rule Violations}
Previous work \cite{ReyesCastro2013, Schluter2018} introduced a weighted transition for rule violation to the automaton. During planning, each rule automaton is evaluated over the word $\LTLword$, and the weight equals the penalty for the respective rule.
Specifically, a self-looping automaton is used in \cite{ReyesCastro2013}, i.e., a weighted transition to the same state is added for a violating transition. However, this formulation does not allow to model violations that cause a penalty once vs. violations that accumulate penalties over time. To account for this, \citet{Schluter2018} extended the work of \cite{ReyesCastro2013}, defining an explicit violation symbol.

Although violations of safety properties can be detected on partial traces, a safety property $\Global \LTLFormulafinite$ will never be satisfied again after being violated. Consequently, violating a safety property leads to a non-accepting state, that only has a self-loop. The automaton thus cannot be brought back to an accepting state anymore, once it has registered a violation.
To obtain penalties for consecutive rule violations, such as multiple times overtaking on the right \cite{Esterle2019a}, we reset the automaton to its initial state, if a violation occurred.
Formally, we modify $\dfa = (\AutomatonStateSet, \AutomatonInitialState, \AutomatonAlphabet, \AutomatonTransitionFunction, \AutomatonAcceptingStates)$ to $\overline{\dfa} = (\AutomatonStateSet, \AutomatonInitialState, \AutomatonAlphabet, \overline{\AutomatonTransitionFunction}, \AutomatonAcceptingStates)$, where
\begin{equation}
	\overline{\AutomatonTransitionFunction}(\AutomatonState, \LTLsymbol) := 
	\begin{cases}
		\AutomatonInitialState & \text{if } \forall \sigma' \in \AutomatonAlphabet: \AutomatonTransitionFunction(\AutomatonTransitionFunction(\AutomatonState, \LTLsymbol), \sigma') = \AutomatonTransitionFunction(\AutomatonState, \LTLsymbol) \land \\ & \AutomatonTransitionFunction(\AutomatonState, \LTLsymbol) \notin \AutomatonAcceptingStates\\
		\AutomatonState' \in \AutomatonAcceptingStates & \text{if } \AutomatonState = \AutomatonInitialState \land \LTLfalive \notin \LTLsymbol \\
		\AutomatonTransitionFunction(\AutomatonState, \LTLsymbol) & \text{else}
	\end{cases}
\end{equation}
Extending our formulation to be invariant to the step size will be the subject of future work.

Let a given rule formalized in LTL $\LTLformula$ be represented by $\overline\dfa_\LTLformula$. The weighting function $\AutomatonWeightingFunction : \AutomatonStateSet \times \AutomatonAlphabet \rightarrow \real$ defines the penalty for violating that rule, assigning each transition a scalar weight:
\begin{equation}
\AutomatonWeightingFunction (\AutomatonState, \LTLsymbol) := \begin{cases}
\weight & \text{if } \forall \sigma' \in \AutomatonAlphabet: \AutomatonTransitionFunction(\AutomatonTransitionFunction(\AutomatonState, \LTLsymbol), \sigma') = \AutomatonTransitionFunction(\AutomatonState, \LTLsymbol) \land \\ & \AutomatonTransitionFunction(\AutomatonState, \LTLsymbol) \notin \AutomatonAcceptingStates \textit{ // safety}\\
\weight & \text{if } \LTLfalive \notin \AutomatonAlphabet \land \AutomatonState \notin \AutomatonAcceptingStates \textit{ // guarantee}\\ 
0 & \text{else}
\end{cases}
\end{equation}
A penalty $\weight$ is returned, if the modified automaton $\overline\dfa_\LTLformula$ gets reset to its initial state (violation of a safety property $\Global \LTLFormulafinite$) or the automaton does not halt in an accepting state (violation of a guarantee property $\Final \LTLFormulafinite$).
Finally, the reward for violating $\LTLformula$ is then defined as
\begin{equation}
\reward(\MarkovState^{k}, \MarkovState^{k+1}) = \AutomatonWeightingFunction (\AutomatonState, \LabeledMarkovLabelingFunction(\MarkovState^{k+1}))
\end{equation}

\subsection{Multi-Objective Reward Function with Priorities}

To model the multi-objective reward (i.e., safety, legal, comfort) and to prevent weight-tuning for a scalar reward, we treat the reward elements as a vector:

\begin{equation}
\rewardVector = 
\begin{pmatrix}
\reward_1 & \hdots & \reward_n
\end{pmatrix}^\transpose
\end{equation}

If goals are meant to be traded off among each other, we perform weighted sum scalarization for these. While \citet{Wang2012} try to find elements on the Pareto-optimal front, the multi-objective formulation for us simplifies, as our objective vector is ordered according to the priority levels. An entry at index $i$ in the reward vector denotes a higher priority than at index $i+1$.

\subsubsection{UCT for vectorized rewards} To incorporate a vectorized reward to the MCTS, we modify the MCTS selection strategy by computing the UCT value per reward vector element. We then compare the vector elements according to the lexicographical order.

\subsubsection{Relaxed lexicographical order}
If employing strict lexicographical order, the selection would favor tree branches that are suboptimal in terms of all other criteria over tree branches, where one single outcome was a collision (if that is the top-level priority). To account for the inaccuracy of the sampling-based approximation of the optimal solution, we implement a relaxation of the lexicographical order, called \textit{Thresholded Lexicographical Ordering} (TLO) \cite{Vamplew2011}. Comparing two reward vectors $\vect{\reward}$, $\vect{\reward}'$ according to TLO, is defined as 
\begin{equation}
	\begin{gathered}
		\vect{\reward} \letlo \vect{\reward}' \iff \\ \exists i : \reward_i \leq \reward'_i \land \forall j < i : (\reward_j > \thesholdtlo_j \land \reward'_j > \thesholdtlo_j ) \lor \reward_j = \reward'_j,
	\end{gathered}
\end{equation}
where $\letlo$ denotes the thresholded lexicographic comparison.
TLO uses a threshold vector $\vect{\thesholdtlo}$ to determine if two goals are sufficiently close so that the next lower priority level can be considered.

\section{Evaluation}
\label{sec:evaluation}
\subsection{Experimental Setup}

We study our approach using the open-source benchmarking and development framework BARK proposed in \cite{Bernhard2020}. We use Spot \cite{Duret-Lutz2016}, a library for model checking, to translate the formalized LTL formula to a DFA, and to manipulate the automata. We implement the runtime monitoring and the multi-objective reward function on top of a template-based MCTS library \cite{Bernhard2019b}. 
\subsubsection{Benchmarking Framework}
BARK is a multi-agent environment tailored to develop interactive behavior models. It allows to easily exchange behavior models for planning, prediction and simulation. BARK already offers a range of behavior models, which we can use for predicting and simulating the other agents. It provides efficient collision checking and realistic maps. A kinematic bicycle model is employed to simulate the vehicles.
\subsubsection{Variants}

We will study 
\begin{itemize}
	\item \textbf{SA} as a scalar single agent variant of MCTS with penalties for collision and comfort,
	\item \textbf{SA-Lex} as a lexicographic baseline implementing the same rewards as SA, but as a vectorized reward,
	\item \textbf{SA-Lex (Zip)} extending SA-Lex by including a penalty for the zipper merge,
	\item \textbf{SA-Lex (SD)} extending SA-Lex by including a penalty for the safe distance (SD) rule.
	\item \textbf{SA-Lex (Zip $>$ SD)} and \textbf{SA-Lex (SD $>$ Zip)} extending SA-Lex by including a penalty for for the zipper merge and the safe distance rule.
\end{itemize}

We define a base reward
\begin{equation}
	\reward_\SubscriptBase = \reward_{a} + \reward_{lat} + \reward_{\Delta v} + \reward_\phi, 
\end{equation}
which consists of penalties for
\begin{itemize}
\item Longitudinal acceleration $\reward_{a} = -w_{a} a^2 \Delta t$
\item Lateral acceleration $\reward_{\mathrm{lat}} = -w_{\mathrm{lat}} |\dot{\theta}| v^2 \Delta t$
\item Difference to desired velocity $\reward_{\Delta v} = -w_v |v-v_{r}| \Delta t$
\end{itemize}
with respective weights $w_\square$, velocity $v$, reference velocity $v_r$, acceleration $a$, orientation rate $\dot{\theta}$ and time increment $\Delta t$.
The convergence of MCTS can be accelerated by incorporating additional domain knowledge. We define a potential function
$\phi(s) = -w_\phi |v - v_r| \Delta t$ and the potential-based shaping function as
\begin{equation}
	\reward_\phi = \DiscountFactor \phi(s_{k+1}) - \phi(s_k)
\end{equation}

$\reward_\SubscriptCol$ denotes the penalty for not colliding. $\reward_\SubscriptSD$ and $\reward_\SubscriptZip$ denote the penalty for violating the safe distance and the zipper merge rules, respectively. The \blockquote{zipper merge} rule requires vehicles, that are on a continuing lane, to let vehicles on an ending lane merge in a zipper fashion. The \blockquote{safe distance} rule requires to leave a safe distance to the vehicle in front. Both rules are defined in \cite{Esterle2020b}.
\refTable{tab:reward_variants} shows the reward vectors of these variants.

\begin{table}[t]
	\scriptsize
	\sisetup{per-mode=symbol}
	\vspace{0.15cm}
	\caption{Variants of $\behaviorModelUnterTest$ showing the reward vector $\rewardVector$ and the size of the reward vector.}
	\centering
	\setlength{\tabcolsep}{5pt}
	\begin{tabular}{lll}
		\toprule
		$\behaviorModelUnterTest$ & Reward vector $\rewardVector$ & size($\rewardVector$)\\
		\midrule
		SA & $\rewardVector = \begin{pmatrix} \reward_\SubscriptCol + \reward_\SubscriptBase \end{pmatrix}^\transpose$ & 1\\
		SA-Lex & $\rewardVector = \begin{pmatrix} \reward_\SubscriptCol & \reward_\SubscriptBase \end{pmatrix}^\transpose$ & 2\\
		SA-Lex (Zip) & $\rewardVector = \begin{pmatrix} \reward_\SubscriptCol & \reward_\SubscriptZip & \reward_\SubscriptBase \end{pmatrix}^\transpose$ & 3\\
		SA-Lex (SD) & $\rewardVector = \begin{pmatrix} \reward_\SubscriptCol & \reward_\SubscriptSD & \reward_\SubscriptBase \end{pmatrix}^\transpose$ & 3\\
		SA-Lex (Zip $>$ SD) & $\rewardVector = \begin{pmatrix} \reward_\SubscriptCol & \reward_\SubscriptZip & \reward_\SubscriptSD & \reward_\SubscriptBase \end{pmatrix}^\transpose$ & 4\\
		SA-Lex (SD $>$ Zip) & $\rewardVector = \begin{pmatrix} \reward_\SubscriptCol & \reward_\SubscriptSD & \reward_\SubscriptZip & \reward_\SubscriptBase \end{pmatrix}^\transpose$ & 4\\
		\bottomrule
	\end{tabular} 
	\label{tab:reward_variants}
	\vspace{-0.15cm}
\end{table}

\subsubsection{Action Space}
We model the discrete action space for the ego agent as \blockquote{lane keeping at constant acceleration} for $\Action \in$ $\{\num{0} \si{\m\per\square\second}, \num{1} \si{\m\per\square\second}, \num{-2} \si{\m\per\square\second}, \num{-8} \si{\m\per\square\second}\}$, \blockquote{lane changing at constant velocity}, and \blockquote{gap keeping} based on the Intelligent Driver Model (IDM) \cite{Treiber2000}. The parameters for the \blockquote{gap keeping} primitive are shown in \refTable{tab:mobil_idm_parameters} with the exception of the desired speed, which we set to \SI{14}{\meter\per\second}.

\subsubsection{Scenarios}
We want to study our approach using real-world data from the INTERACTION dataset \cite{Zhan2019}. However, we cannot just replace the agent with our model and replay the other agents, as the other agents would not react to our model under test anymore. We thus preserve the initial configuration of the vehicles from the dataset but simulate them according to $\behaviorOthers$.
Each vehicle is simulated as the ego agent in one scenario. 
The ego agent is controlled by the behavior model $\behaviorModelUnterTest$. The scenario is passed successfully, if the ego agent reaches its goal region, which we create from the last pose of the agent in the dataset.

\subsubsection{Behavior Model for Others}
The actions of the other vehicles are calculated using the behavior model $\behaviorOthers$, for which we employ a rule-based model {\classnamestyle{BehaviorMobilRuleBased}}, with IDM as a longitudinal and MOBIL \cite{Kesting2015} as a lateral model, and a lane filtering mechanism on top of MOBIL. The model is available as open-source in BARK. \refTable{tab:mobil_idm_parameters} shows the parameters. To cause challenging situations for the ego vehicle, we let the other vehicles travel at a desired speed of \SI{10}{\meter\per\second}, and the ego vehicle at \SI{14}{\meter\per\second}.

\subsubsection{Rule Evaluators}
\begin{figure}[tb]
	\vspace{0.15cm}
	\centering
	\scriptsize
	\def\svgwidth{\columnwidth}
	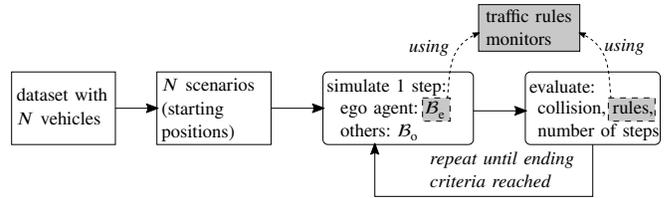
	\caption{Framework to evaluate the behavior of the ego agent $\behaviorModelUnterTest$ in a closed-loop simulation. The initial starting positions for the vehicles are taken from the dataset. The other vehicles are simulated using a behavior model $\behaviorOthers$. The traffic rule monitors are used within $\behaviorModelUnterTest$ and to evaluate the simulation.}
	\label{fig:evaluation_toolchain}
	\vspace{-0.15cm}
\end{figure}

BARK provides an abstract evaluator class, that calculates a given metric such as collision or step count based on the simulated world state. We have extended this evaluator concept to evaluate arbitrary LTL formulas on finite traces, and made it available within BARK. Each rule is captured in a rule monitor, which we can use to monitor compliance throughout the simulation. Both the behavior model and the evaluator employ the same rule monitor, which we have contributed as open-source\footnote{\url{https://github.com/bark-simulator/rule-monitoring}}. This allows us to study whether $\behaviorModelUnterTest$ truly satisfies the modeled rules. 
\refFigure{fig:evaluation_toolchain} shows the resulting evaluation framework.

\subsection{Quantitative Evaluation}
\begin{figure*}[tb]
	\vspace{0.15cm}
	\centering
	\input{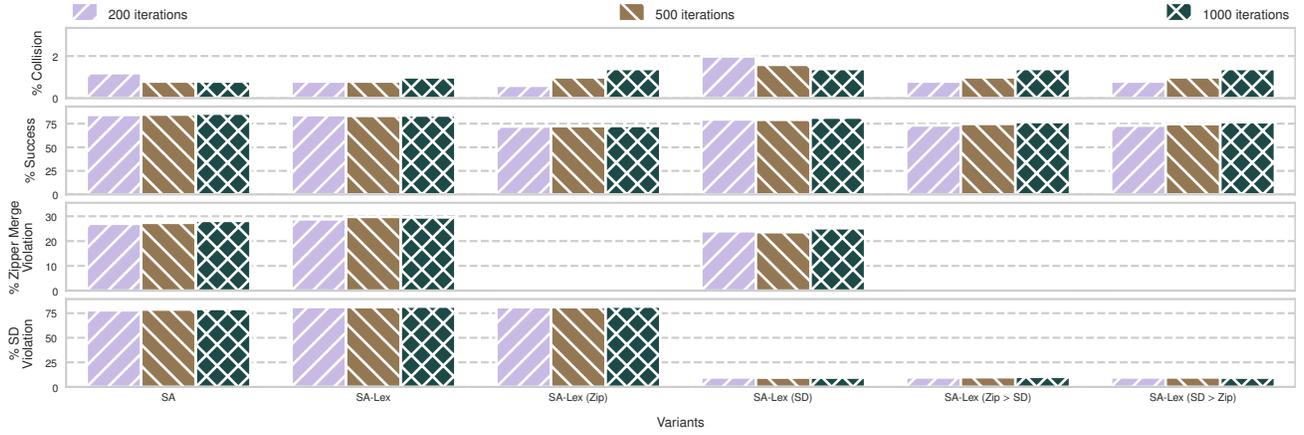}  
	\caption{Benchmark of the variants for $\behaviorModelUnterTest$. The comparison of \textit{SA-Lex (Zip > SD)} and \textit{SA-Lex (SD > Zip)} with reversed priorities shows that both modeled rules can be satisfied at the same time and thus no tradeoff between those rules is required for our scenario.}
	\label{fig:benchmark_results}
	\vspace{-0.15cm}
\end{figure*}

We evaluate $\behaviorModelUnterTest$ for \num{200}, \num{500} and \num{1000} search iterations.
\refFigure{fig:benchmark_results} shows the share of controlled agents to collide, successfully reach the goal region (within \SI{30}{\second}) and to violate the zipper merge or safe distance rule. 

We observe nearly no collisions, which indicates that MCTS with high level actions can approximate the solution well, if predicted and simulated behavior are matching. 
\textit{SA}, \textit{SA-Lex} and \textit{SA-Lex (SD)} do violate the zipper merge at about \num{25}\%, which shows the this rule will not be satisfied implicitly and motivates its explicit modeling. \textit{SA-Lex (SD)} violates it to some less extent, as keeping a safe distance all the time sometimes leaves enough space for the merging vehicle to fit in.  
The variants not implementing the safe distance rule usually do not leave much space to the front vehicle, as the correct prediction model yields an accurate estimate of what will happen given a certain action, as long as the tree is explored enough. However, in order to obey to the safe distance rule, and to be prone to unexpected actions of the human drivers, it must be modeled within the planner. The remaining safety distance violations stem from unsafe initial states at the beginning of the scenario. 
The variants \textit{SA-Lex (Zip > SD)} and \textit{SA-Lex (SD > Zip)} do not cause any violations to the zipping rule, but also keep a safe distance for most scenarios. The results for both are similar, which indicates that there exists no trade-off in our scenarios between the safe distance and the zipper rule.

When varying the number of search iterations, the results remain mostly unchanged for the variants \textit{SA}, \textit{SA-Lex} and \textit{SA-Lex (SD)}. 
With more search iterations, $\behaviorModelUnterTest$ becomes more certain about which actions will prevent a zipper merge violation. As $\behaviorOthers$ is not necessarily collision-free (it does not employ any prediction), this creates more dense situations, and thus explains the slightly increasing number of collisions.

\begin{table}[tb]
	\scriptsize
	\sisetup{per-mode=symbol}
	\caption{Parameters of {\footnotesize{\textsf{BehaviorMobilRuleBased}}} for $\behaviorOthers$. The lane change rules serve as an additional filter of the lanes, to which MOBIL can choose to change to.}
	\centering
	\setlength{\tabcolsep}{5pt}
	\begin{tabular}{llll}
		\toprule
		& Parameter & Unit & Value \\ 
		\midrule
		\multirow{5}{*}{IDM} & Desired velocity & [\si{\meter\per\second}] & 10\\
		& Maximum acceleration & [\si{\meter\per\second\squared}] & 1.7\\
		& Desired time headway & [\si{\second}] & 2.5\\
		& Comfortable deceleration & [\si{\meter\per\second\squared}] & 2\\
		& Minimum distance & [\si{\meter}] & 2\\
		\midrule
		\multirow{3}{*}{Lane change rules} & Min. rear distance & [\si{\meter}] & 0.5\\
		& Min. front distance & [\si{\meter}] & 1\\
		& Time Gap & [\si{\second}] & 0.5\\
		\midrule
		\multirow{3}{*}{MOBIL} & Politeness Factor & [1] & 0\\
		& Safe deceleration & [\si{\meter\per\second\squared}] & 4\\
		& Acceleration threshold & [\si{\meter\per\second\squared}] & 0.2\\
		\bottomrule
	\end{tabular} 
	\label{tab:mobil_idm_parameters}
\end{table}

\section{Conclusion and Future Work}
\label{sec:conclusion}
In this work, we investigated the problem of how interactive behavior planning for an autonomous vehicle can be modeled to obey the traffic rules, and how this can be evaluated and tested.
We modeled the interaction as a dynamic game and used MCTS to solve the problem, while incorporating ideas from model checking and multi-objective optimization.

In our evaluation, we demonstrated the capabilities of our approach in a closed-loop simulation for a merging scenario at dense traffic in a systematic way. For this, we selected two rules that apply to this situation, namely to keep a safe distance and to merge in a zipper fashion. 

Our new method allows us to formalize multi-agent rules, that apply to more than one agent, whose future motion is modeled interactively. With this, multi-agent time-dependent traffic rules can be studied on whether they have been correctly formalized, and on what the ramifications of the rules are on safety and progress. %
Future work should investigate the probabilistic interpretation of predicates.

\section*{Acknowledgment}
This research was funded by the Bavarian Ministry of Economic Affairs, Regional Development and Energy.

\renewcommand{\bibfont}{\footnotesize}
\printbibliography
\end{document}